\documentclass[12pt,a4paper,twoside,twocolumn]{article}
\makeatletter\if@twocolumn\PassOptionsToPackage{switch}{lineno}\else\fi\makeatother

\usepackage{amsfonts,amssymb,amsbsy,latexsym,amsmath,tabulary,graphicx,times,xcolor}
\usepackage[T1]{fontenc}
\usepackage{url,multirow,morefloats,floatflt,cancel,tfrupee}
\makeatletter

\AtBeginDocument{\@ifpackageloaded{textcomp}{}{\usepackage{textcomp}}}
\makeatother
\usepackage{colortbl}
\usepackage{xcolor}
\usepackage{pifont}
\usepackage{amsmath}
\usepackage[nointegrals]{wasysym}
\usepackage[superscript,biblabel]{cite}
\usepackage{bm}
\let\vec\bm

\makeatletter
\@ifundefined{etal}{}{}

\AtBeginDocument{
\expandafter\ifx\csname eqalign\endcsname\relax
\def\eqalign#1{\null\vcenter{\def\\{\cr}\openup\jot\m@th
  \ialign{\strut$\displaystyle{##}$\hfil&$\displaystyle{{}##}$\hfil
      \crcr#1\crcr}}\,}
\fi
}

\@ifundefined{tsGraphicsScaleX}{\gdef\tsGraphicsScaleX{1}}{}
\@ifundefined{tsGraphicsScaleY}{\gdef\tsGraphicsScaleY{.9}}{}
\def\checkGraphicsWidth{\ifdim\Gin@nat@width>\linewidth
	\tsGraphicsScaleX\linewidth\else\Gin@nat@width\fi}

\def\checkGraphicsHeight{\ifdim\Gin@nat@height>.9\textheight
	\tsGraphicsScaleY\textheight\else\Gin@nat@height\fi}

\def\fixFloatSize#1{}
\let\ts@includegraphics\includegraphics

\def\inlinegraphic[#1]#2{{\edef\@tempa{#1}\edef\baseline@shift{\ifx\@tempa\@empty0\else#1\fi}\edef\tempZ{\the\numexpr(\numexpr(\baseline@shift*\f@size/100))}\protect\raisebox{\tempZ pt}{\ts@includegraphics{#2}}}}

\renewcommand{\includegraphics}[1]{\ts@includegraphics[width=\checkGraphicsWidth]{#1}}
\AtBeginDocument{\def\includegraphics{\@ifnextchar[{\ts@includegraphics}{\ts@includegraphics[width=\checkGraphicsWidth,height=\checkGraphicsHeight,keepaspectratio]}}}

\DeclareMathAlphabet{\mathpzc}{OT1}{pzc}{m}{it}

\def\URL#1#2{\@ifundefined{href}{#2}{\href{#1}{#2}}}

\edef\fntEncoding{\f@encoding}

\makeatother

\newif\ifmultipleabstract\multipleabstractfalse

\usepackage[hmargin=1.8cm,vmargin=2.2cm]{geometry}
\makeatletter
\def\author#1{\gdef\@author{\hskip-\dimexpr(\tabcolsep)\hskip1pt\parbox{\dimexpr\textwidth-1pt}{\centering #1}}}

\let\@articletype\@empty \def\articletype#1{\gdef\@articletype{{\fontsize{14}{16}\selectfont #1}}}

\usepackage{fancyhdr}

\fancypagestyle{headings}{\fancyhf{}\fancyhead[RE]{\MakeTextUppercase{\textbf{\RunningAuthor}}}\fancyhead[LE]{\textbf{\thepage}}\fancyhead[LO]{\MakeTextUppercase{\textbf{\RunningHead}}}\fancyhead[RO]{\textbf{\thepage}}}
\fancypagestyle{plain}{\fancyhf{}\fancyfoot[C]{\textbf{\thepage}}}

\AtBeginDocument{\usepackage{lastpage}}
\def\title#1{%
  \gdef\@title{%
    \ifx\@articletype\@empty\else\@articletype~\\\fi%
     #1}%
}

\usepackage{abstract}
\def\abstractname{\textbf{Abstract}}
\renewenvironment{onecolabstract}
{\vspace*{-.4pc}\trivlist\item[]\leftskip1pt\noindent\selectfont\hfill\abstractname\hfill\mbox{\null}\par\ignorespaces}{\endtrivlist}

\def\NormalBaseline{\def\baselinestretch{1.1}}

\usepackage{textcase}
\usepackage[explicit]{titlesec}
\titleformat{\section}[block]{\NormalBaseline\boldmath\bfseries}
{\thesection.}
{6pt}
{#1}
[]
\titleformat{\subsection}[hang]{\NormalBaseline\filright\itshape}
{\thesubsection.}
{6pt}
{#1}
[]
\titleformat{\subsubsection}[runin]{\NormalBaseline\filright\itshape}
{\hspace{16pt}\thesubsubsection}
{6pt}
{#1}
[]
\titleformat{\paragraph}[runin]{\NormalBaseline}
{\theparagraph}
{6pt}
{#1}
[]
\titleformat{\subparagraph}[runin]{\NormalBaseline}
{\thesubparagraph}
{6pt}
{#1}
[]

\titlespacing{\section}{0pt}{1.5\baselineskip}{.2\baselineskip}  
\titlespacing{\subsection}{0pt}{1.5\baselineskip}{.2\baselineskip}  
\titlespacing{\subsubsection}{0pt}{1.5\baselineskip}{.2\baselineskip}  
\titlespacing{\paragraph}{0pt}{.5\baselineskip}{10pt}  
\titlespacing{\subparagraph}{0pt}{.5\baselineskip}{10pt}

\usepackage{caption}
\DeclareCaptionLabelFormat{figlabel}{Figure #2}
\date{}
\makeatother

\usepackage{float}
\usepackage{subcaption}

\begin{document}
 
\title{A Geometric Kinematic Model for Flexible Voxel-Based Robots}

\def\RunningHead{Gaits for voxel robots}
\def\RunningAuthor{Tebyani et al.}
\author{Maryam Tebyani\textsuperscript{1, *},  Alex Spaeth\textsuperscript{1, 2},  Nicholas Cramer\textsuperscript{3}, and Mircea Teodorescu\textsuperscript{1, 2}}
\maketitle

\thanks{
\textbf{1} Department of Electrical and Computer Engineering,
University of California, Santa Cruz, Santa Cruz, California, United States \\
\textbf{2} Genomics Institute,
University of California, Santa Cruz, Santa Cruz, California, United States\\
\textbf{3} NASA Ames Research Center, Moffett Field, California, United States\\
\textbf{*} mtebyani@ucsc.edu
}

%%%%%%%%%%%%%%%%%%%%%%%%%%%%%%%%%%%%%%%%%%%%%%%%%%%%%%%%%%%%%%%%%%%%%%%%%%

{\begin{onecolabstract}

Voxel-based structures provide a modular, mechanically flexible periodic lattice which can be used as a soft robot through internal deformations. To engage these structures for robotic tasks, we use a finite element method to characterize the motion caused by deforming single degrees of freedom and develop a reduced kinematic model. We find that node translations propagate periodically along geometric planes within the lattice, and briefly show that translational modes dominate the energy usage of the actuators. The resulting kinematic model frames the structural deformations in terms of user-defined control and end effector nodes, which further reduces the model size. The derived Planes of Motion (POM) model can be equivalently used for forward and inverse kinematics, as demonstrated by the design of a tripod stable gait for a locomotive voxel robot and validation of the quasi-static model through physical experiments. 

\smallskip\noindent\textbf{Keywords:}
Modular Robots, Flexible Locomotion, Kinematic Modeling   

\end{onecolabstract}}
 
%%%%%%%%%%%%%%%%%%%%%%%%%%%%%%%%%%%%%%%%%%%%%%%%%%%%%%%%%%%%%%%%%%%%%%%%%%
 
\section{Introduction} 

Soft robots leverage the deformation of their body to create intelligent motion. The changes to the state of their body are a function of the robot's material properties and geometry \cite{rus_design_2015}. 

Previously, Jenett et al. proposed a new class of soft continuum robots where cellular composite structures with low density and high specific stiffness form modular soft robots \cite{jenett_digital_2016, jenett_discretely_2020}. Discrete lattice-based building materials can combine large numbers of identical discrete volumetric pixels, or ``voxels,'' to form a metamaterial \cite{rafsanjani_programming_2019} with several promising applications, allowing advances in aircraft where wings can passively and actively morph to increase aerodynamic efficiency and control authority \cite{cramer_elastic_2019}, large-scale spacecraft can be autonomously assembled and monitored in outer space\cite{trinh_robotically_2017,jenett_design_2017}, and modular flexible robots can generate undulatory locomotion using linear actuation \cite{cramer_design_2017}. Such modular robotic structures benefit from reconfigurability, element replacement, scalability, and decreased production costs \cite{brunete_current_2017, siciliano_modular_2016}. 

\begin{figure*}
    \centering
    \includegraphics{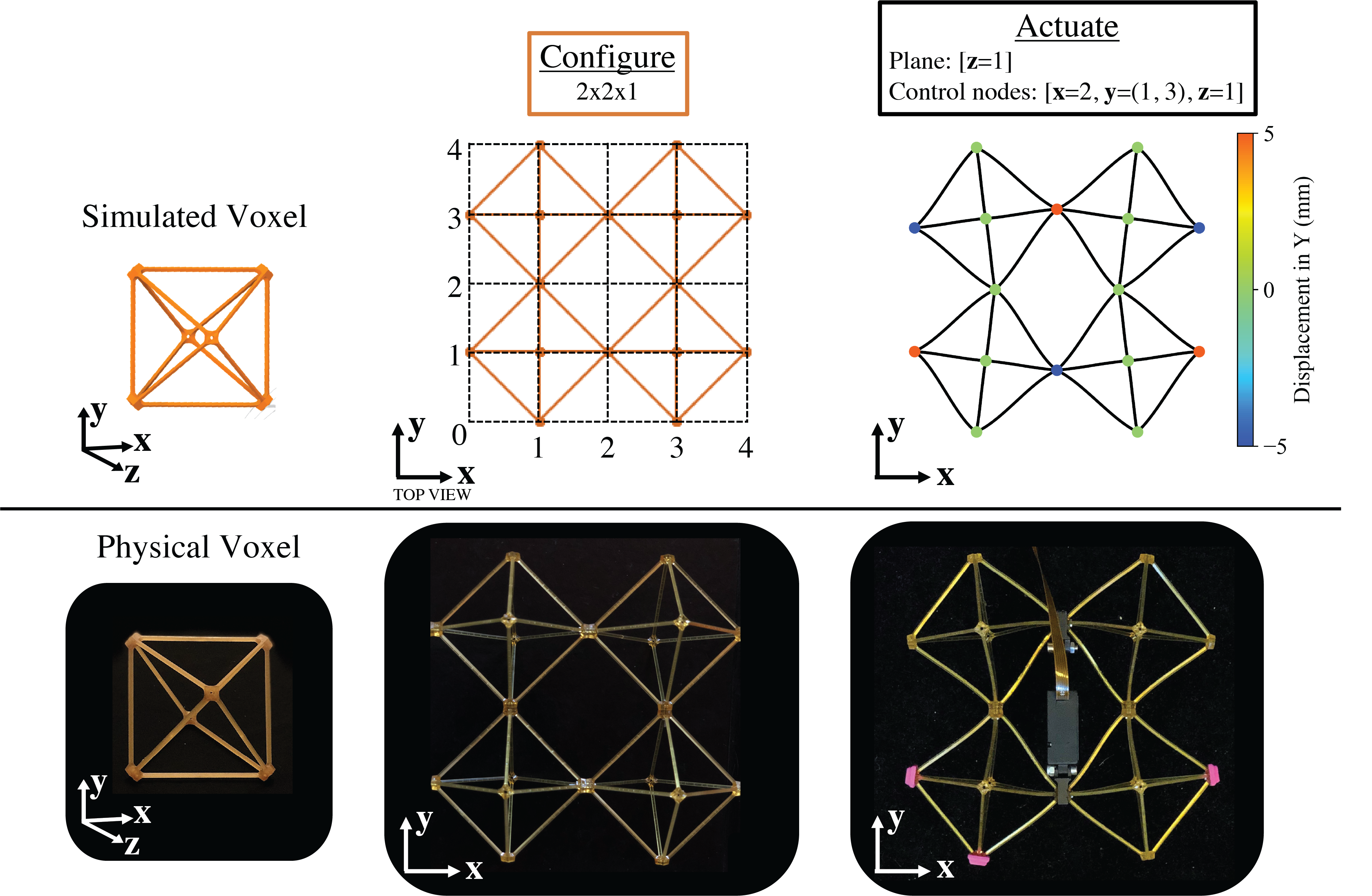}
    \caption{Voxel robots are designed by specifying the configuration and actuation of the voxel structure. The images on the left show a single voxel element as a 3D model on top and its physical counterpart on bottom. The middle images show a structure of 4 voxels.  
    The images on the right show the resulting displacement of the voxel body when the actuation is applied via a linear actuator between two control nodes. Beam bending is approximated using a cubic spline for visualization purposes.}
    \label{fig:fig1}
\end{figure*} 

However, the design and modeling of voxel-based continuum robots is limited by the complexity of available modeling approaches. A review of model-based and model-free static and dynamic controllers for soft robotic manipulators concludes that hybrid modeling is a promising area of research \cite{george_thuruthel_control_2018}. Recently, machine learning has been employed to develop differential models for closed-loop soft robotic control \cite{chin_machine_2020}. Model-based methods to predict the movement of soft robots often segment the body into constituent elements which range in computational cost from finite element models, which can take hours to simulate, to piecewise constant curvature models \cite{webster_design_2010, katzschmann_dynamic_2019, kuppuswamy_learning_2011} which can be used for real-time control.

Similar methods have been used to characterize a beam-shaped robotic swimmer constructed from voxels \cite{jenett_discrete_2020}. However, more general robots cannot be modeled this way because of their three-dimensional lattice structure. Voxel structures have been characterized as materials \cite{rafsanjani_programming_2019, popescu_vibrations_2019}, but limited work has been done to show how these structures can be employed for robotic tasks. 
Previously, Cramer et al. modeled quasi-static locomotion of a voxel robot using a linear finite element method, but finite element approaches are not suitable for real-time applications \cite{cramer_design_2017}.
 
In this paper, we attempt to address this deficiency by introducing a simple geometric model of deformation in voxel structures which enables a computationally cheap approximation of the position of nodes of interest as a function of actuator-induced structural deformation. This model can be used for the design of locomotive gaits, as well as for embedded control with modest resource requirements, highlighting how actuation of compliant structures can produce constrained motion which still benefits from the structure's embedded intelligence. 
First, the motion produced by actuating voxel structures is studied. Emergent patterns in the voxel movement are used to define general relations for the motion of nodes under linear actuation, allowing the development of a geometric model which we call the Planes of Motion (POM) model. Next, a voxel-based quadrupedal robot is designed using the derived simulation. 
 
Experimental results show that the simplified model produces a kinematic template of voxel node displacements.   
 
\section{Methodology} \label{s:methodology}

The voxels used in this work are regular cubic octahedra structured as 12 beams joined at 6 nodes. This geometry can be understood as three orthogonally intersecting squares. Voxels are assembled into flexible structures which can be deformed by embedded linear actuators. 

To illustrate the deformation behavior of our voxel structures, we use the open-source Python library PFEA \cite{cramer_design_2017} to simulate the displacements of individual nodes. PFEA is a finite-element solver optimized for voxel structures, which calculates node positions under the assumption that each node has zero extent and all beams are linear. Additionally, both PFEA and our geometric model to be introduced later make a quasi-static assumption; the low inertia of the voxels and high viscous friction of the actuators result in a slow-moving system which remains close to equilibrium.

PFEA results are compared to physical voxels in figure \ref{fig:fig1}. Four voxels are arranged in a two-by-two square formation, and a linear actuator is placed between them. As the actuator expands, it forces the positions of opposite nodes apart, deforming the full structure.  In the figure, nodes are colored according to their displacement in the vertical direction as calculated by PFEA.

\begin{figure*}
    \centering
    \includegraphics[width=\linewidth]{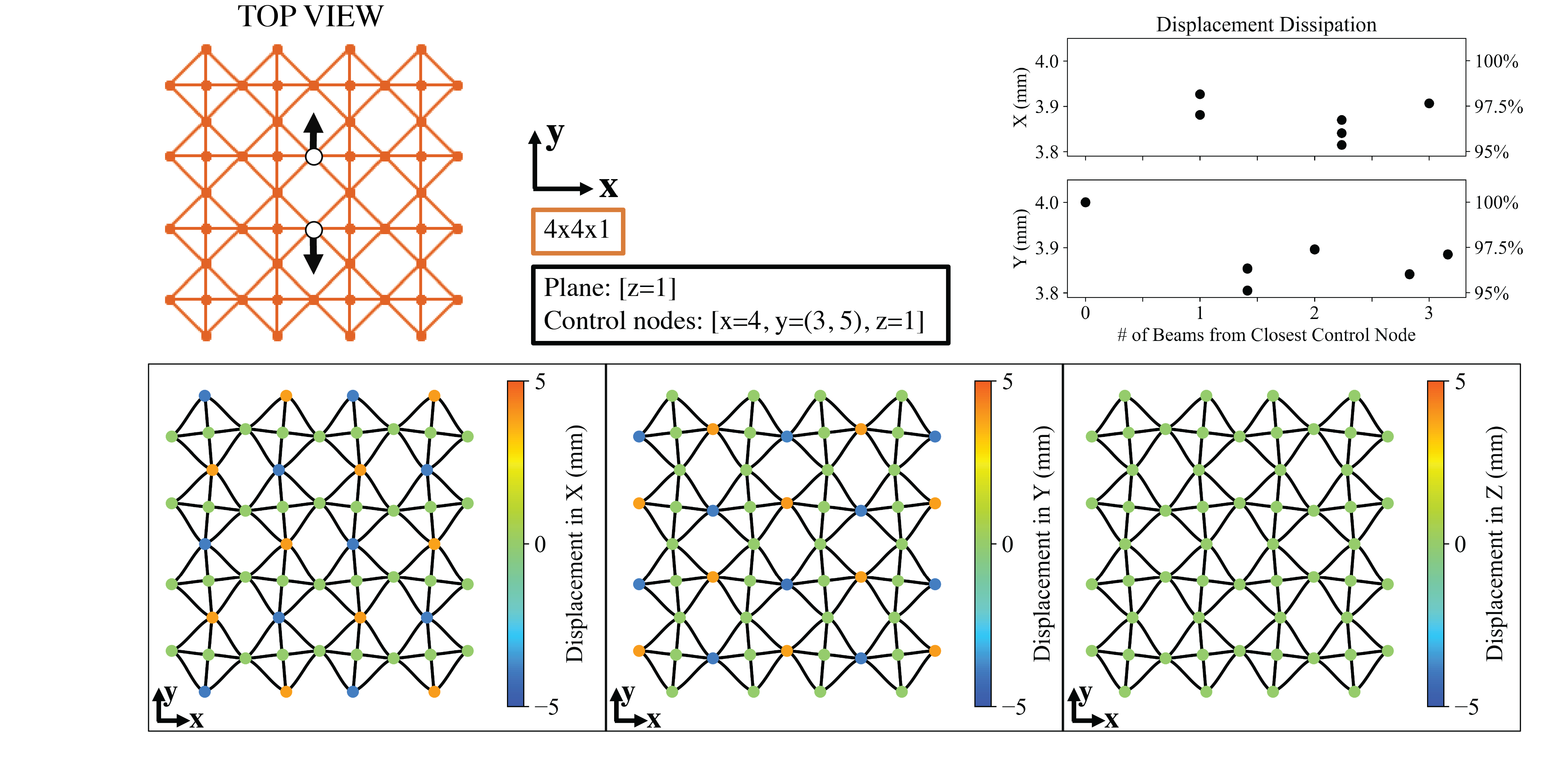}
    \caption{A 4×4×1 structure is actuated to demonstrate the emergent displacement pattern arising from the constrained degrees of freedom. The node coloring of the left bottom image shows the displacement magnitudes in the X dimension, the middle shows the displacement magnitudes in the Y, and the right in the Z. Finally, the Displacement Dissipation plot in the top right shows how energy absorbed by other modes, like node rotation and beam bending, diminishes node displacements.}
    \label{fig:fig2}
\end{figure*}

\begin{figure*}
    \centering
    \includegraphics{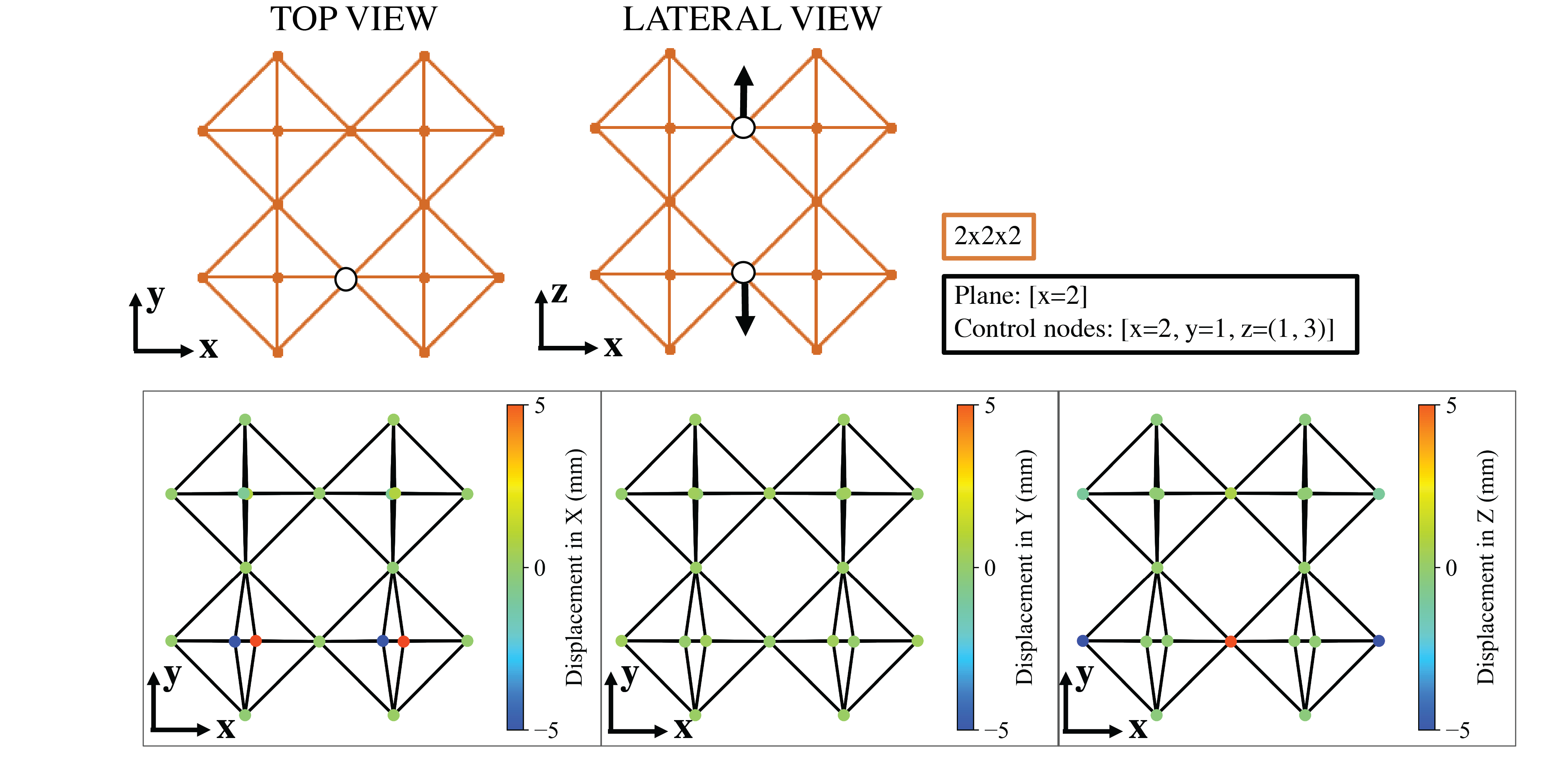}
    \caption{A 2×2×1 voxel configuration is actuated to demonstrate the minimal displacement of neighboring planes. Beams are plotted here as straight lines to make the three-dimensional structure clearer.}
    \label{fig:fig3}
\end{figure*}

\subsection{Geometric Relations}

Studying the configuration and actuation shown in figure \ref{fig:fig1}, it is notable that constraints and loading conditions in simulation and the actuator mounts on the physical structure inhibit the rotational degrees of freedom of the nodes that they connect; the actuation strictly displaces a single, linear degree of freedom of the selected pair of control nodes. 

For example, the actuator in figure \ref{fig:fig1} connects neighboring nodes which share an $x$ and $z$ coordinate. Voxels are made of narrow plastic beams which can bend easily but retain their shape and length, so lattice deformations effectively rotate voxels in place. This displaces other nodes which fall on the $z=1$ plane, in a direction that accommodates the inciting node motion. Thus, a displacement pattern emerges and propagates throughout the actuated plane. This pattern is shown in figure \ref{fig:fig2}, where the loads and constraints from figure \ref{fig:fig1} are applied on a larger voxel structure. Although the same force is applied, the displacements are smaller in figure \ref{fig:fig2} because more voxels must be deformed. This pattern does not decay with distance, as demonstrated by the Displacement Dissipation plot in figure \ref{fig:fig2}. This implies that node translation dominates energy usage, as only up to 4.9\% of displacement is lost to other modes, such as node rotation or beam bending.

Figure \ref{fig:fig3} further demonstrates that significant displacement within the structure is constrained to specific nodes as a function of the voxel geometry. Nodes outside of the actuated plane do not displace translationally. The rotation of the voxel within a plane does rotate its out-of-plane nodes, but this rotation seems to dissipate in the beams without rotating the next layer of voxels. For instance, for this actuator in the $z=1$ plane, other nodes on the $z=0$ and $z=2$ planes will rotate to remain in line with their voxels, but without leading to any movement of the adjacent nodes on the $z=-1$ and $z=3$ planes, which would correspond to the rotation of other voxel layers.

With this in mind, the observation underlying our model is that in a voxel lattice, deformations amounting to a rotation of one of the squares making up each voxel require the least energy to perform, and propagate through the lattice within a single plane by rotating the corresponding square in the opposite direction in each adjacent voxel. The result is that nodes which lie on the same plane experience displacements of the same magnitude, with direction alternating along each axis.

\begin{figure}
    \centering
    \includegraphics[width=\linewidth]{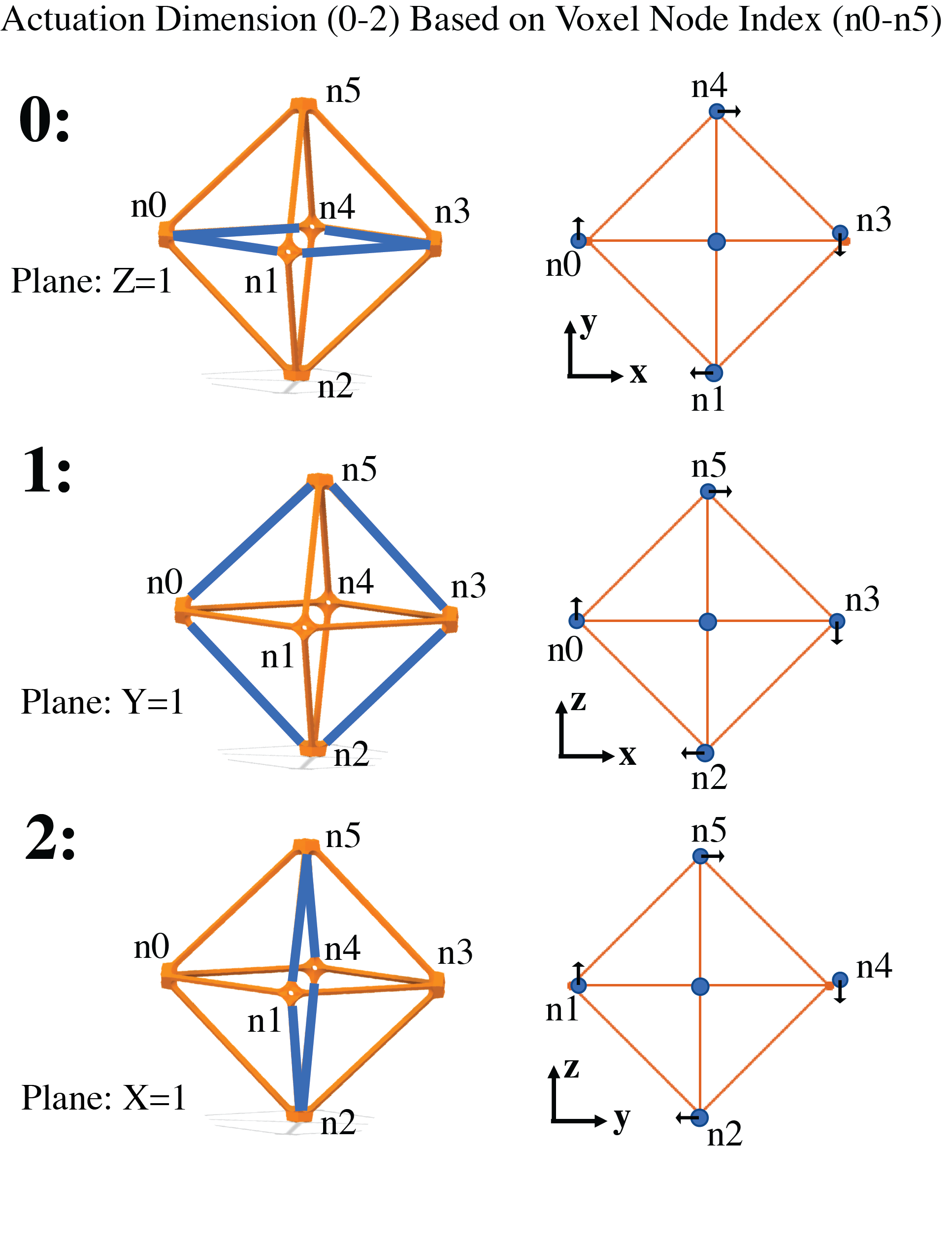}
    \caption{The numerical convention for voxel node and actuation dimension indexing. Actuation along a given plane, highlighted in blue (left), will translate voxel nodes in that plane about the corresponding voxel center, as indicated with the black arrows (right).}
    \label{fig:fig4}
\end{figure}

Figure \ref{fig:fig4} illustrates this alternation as well as our numerical conventions for numbering of voxel nodes and actuation dimensions. Each actuation dimension rotates a corresponding square of the voxel, highlighted in blue on the voxel image on the left. This leads to displacements of the nodes located on that square, shown with arrows emanating from those nodes on the right.
To approximate how voxel nodes will move based on the plane of motion that is activated by the actuated nodes, we consolidate this information into three matrices $A_d$ corresponding to the three actuation directions $d$, and indexed by actuation dimension and voxel node index, as follows:

\begin{align*}
A_0 &= \begin{pmatrix}
0 & 0 & 0 & 0 & 0 & 0 \\
0 & 0 & 1 & 0 & 0 & -1 \\ 
0 & -1  & 0 & 0 & 1 &  0 \\
\end{pmatrix}\\
A_1 &= \begin{pmatrix}
0 & 0 & 1 & 0 & 0 & -1 \\
0 & 0 & 0 & 0 & 0 & 0 \\ 
-1 & 0 & 0 & 1 & 0 & 0 \\
\end{pmatrix}\\
A_2 &= \begin{pmatrix}
0 & -1 & 0 & 0 & 1 & 0 \\
1 & 0 & 0 & -1 & 0 & 0 \\ 
0 & 0 & 0 & 0 & 0 & 0 \\
\end{pmatrix}
\end{align*}

The columns of $A_d$ specify how a control node displacing along direction $d$ will affect the position of each node of any voxels in the corresponding plane of motion, with direction and node indices according to the convention of figure \ref{fig:fig4}. In the model, it is useful to refer to these columns as vectors in space, for which we introduce the notation $\vec a_{d,n}$ to refer to the $n$th column of the matrix $A_d$, i.e. the displacement produced on node $n$ of each voxel in the implicated plane by a unit control node displacement in direction $d$.

\subsection{Planes of Motion (POM) Model}
 
The arrays $A_d$ are then used to define a reduced-dimension kinematic model of actuated voxel structures by considering a subset of the voxel nodes. 
At time $t$, the 3D displacement of the end effector nodes is defined by the current displacement of the control nodes. The vector $\vec q(t)$ in configuration space describes the displacement of the control nodes, and $\vec x_j(t)$ is the 3D displacement of the $j$th end effector. 
The connectivity of the voxel structure is defined by the array $c_{ij}$, which specifies the effect of control node $i$ on end effector $j$. These entries are zero if the two nodes do not share a plane of motion; otherwise, because of the alternating nodal displacement pattern, they are set to -1 or +1 depending on the parity of the position difference. Together with the actuation dimension $d_i$ of the $i$th control node and the node index $n_j$ defining which node of its corresponding voxel the end effector occupies, this map fully specifies how displacements of a given control node deform the voxel structure. These geometric constants, specified numerically according to the conventions of figure \ref{fig:fig4}, are calculated in software based on more physically intuitive quantities: the grid position of the voxel and corresponding within-voxel node ID of each control or end effector node, and the actuation direction of any control nodes.

Finally, the kinematic motion is simulated by calculating the contributions of each control node to each end effector node's position. So long as no two control nodes share a plane of motion, they will specify orthogonal displacements for any affected end effectors, which can simply be summed. For each control node $i$, a column $\vec a_{d_i,n_j}$ corresponding to node index $n_j$ is selected from the actuation matrix $A_{d_i}$ for the actuation dimension $d_i$. The result is multiplied by the masked control node displacement, then summed for all control nodes.

\begin{equation} \label{e:pom}
     \vec x_j(t)  = \sum_{i=0}^{M-1} (\vec a_{d_i,n_j}) (c_{ij} q_i(t))
\end{equation}

In summary, we reduce the space of nodes to focus on the nodes of interest: the actuated nodes, and those intended to interact with the environment. It is important to note that the POM model defines the relationship between two sets of node displacements, regardless of their robotic function. So far, we have designated a control node which is connected directly to the actuator and analyzed the movement of chosen ``end effectors''. Instead, we could just as well have defined a desired displacement for an end effector node and computed the corresponding actuator displacement. Essentially, the forward and inverse kinematic models of the robot are derived in the same way and take the same form. In fact, the same simulation is used to compute both the forward and inverse kinematics for the robotic locomotor presented in the next section.

\section{Results}

\begin{figure*}
    \centering
    \includegraphics{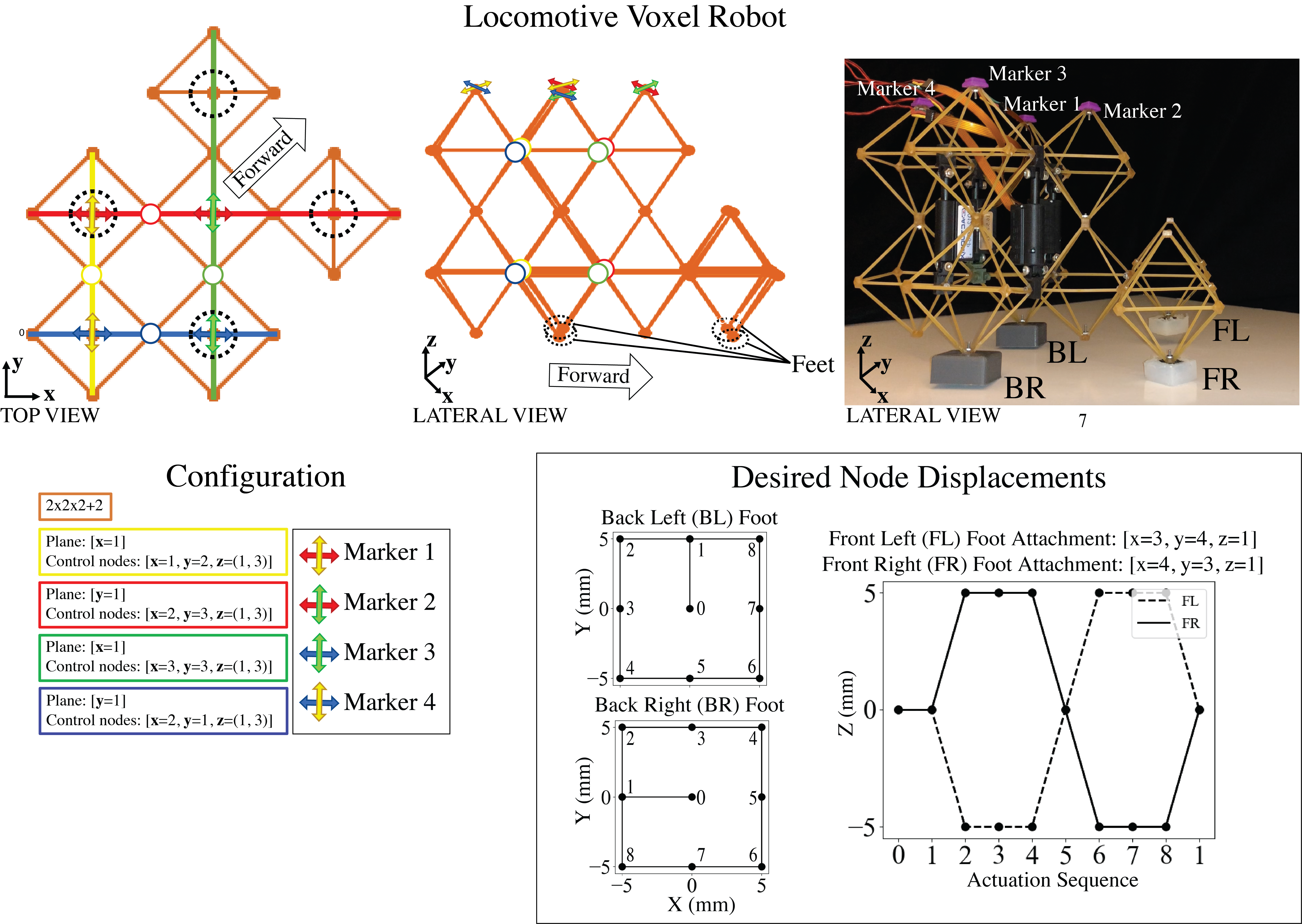}
    \caption{Top: The voxel configuration of the proposed locomotive robot. The two left schematics show a top view and a lateral view of the configuration, where 4 actuators are depicted with colored dots, 4 feet nodes are indicated by black dotted circles, and 4 markers are indicated by crossing arrows. The right image shows the physical prototype with the 4 physical actuators, 4 feet, and 4 (purple) markers. Bottom: The voxel configuration and actuator placement is detailed using the previously defined global coordinate system and a legend is given for the 4 markers shown in the schematics. On the right, the desired node displacements of the 4 feet to achieve the proposed locomotive gait are shown. The back feet nodes travel along the ground in the $x$ and $y$ direction while the front feet nodes travel along the vertical $z$ dimension. These node displacements and robot configuration are fed into the POM model to produce the actuator positions required to achieve the feet displacements shown.}
    \label{fig:fig5}
\end{figure*}

Previous work showed that rectilinear locomotion can be achieved by periodically deforming a voxel structure using bi-directional friction to promote forward motion \cite{cramer_design_2017}. The feet of these robots slid along the ground, but real walking gaits are generated by lifting and lowering feet.
As a step towards this type of gait, we designed an intermediate locomotive robot where the front feet raise and lower while the rear feet play a supporting role, remaining planted on the ground while shifting back and forth to adjust the robot's balance. The core of this robot consists of the 2×2×2 voxel structure shown in figure \ref{fig:fig3}, to which two additional voxels with liftable front feet are attached; this configuration is detailed in figure \ref{fig:fig5}.

This robotic configuration was previously used to demonstrate how spiking neural state machines can construct central pattern generators and account for the voxel's elastic properties through proprioceptive feedback and frequency entrainment \cite{spaeth_neuromorphic_2020, spaeth_spiking_2020}. Here, we provide the geometric intuition and modeling required to develop the desired gait and foot fall pattern, which highlight the benefits of reduced state kinematic models of flexible robots.  

\begin{figure*}
    \centering
    \includegraphics{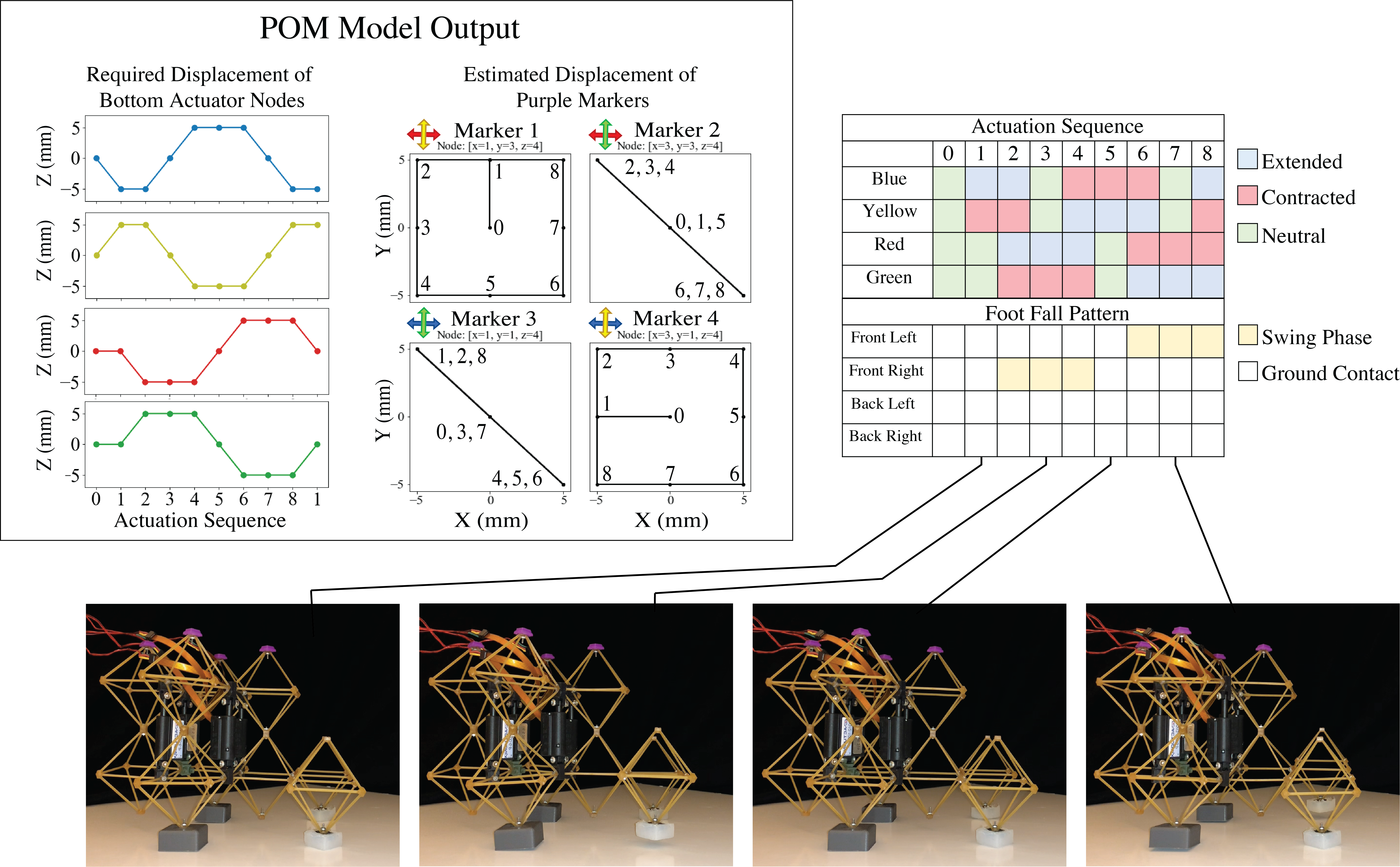}
    \caption{The POM Model Output plots show the required node displacements of the bottom actuator connection sites to achieve the desired gait, and the expected displacement of the purple markers obtained from plugging in the voxel configuration and desired feet motions to the POM model. This output defines the actuation sequence as a function of the foot fall pattern. The bottom row shows the physical robot in key stages throughout the sequence.}
    \label{fig:fig6}
\end{figure*}

\subsection{Gait Design}  

The robot is controlled by four linear actuators placed vertically between the bottom and top layer of its core. Each actuator is connected between two nodes in the voxel structure, shown as colored circles in figure \ref{fig:fig5}; the associated POM are indicated with colored lines. In a forward kinematic view, where actuators are defined as control nodes, voxel nodes which fall on these planes can be controlled along a single dimension according to the general matrix relations given in Section \ref{s:methodology}. For instance, the red actuator controls nodes that fall along the red line in either the $x$ or $z$ dimension, depending on the node ID.

Nodes which fall on the intersection of two actuators' POM can be actuated independently in two directions. Four such nodes are designated as end effectors and indicated by purple markers in the figure.
The four feet of the robot are indicated by the black dotted circles. The two back feet move identically to markers 1 and 4 due to the symmetry of the robot.
For example, to move the back left foot ``forward'', the red actuator needs to displace the end effector in positive $x$, and the yellow actuator needs to displace the end effector in positive $y$. 
The motion of the front feet is complicated by the irregularity of the voxel structure; the POM model requires at least two antagonistic voxels. Instead, the front voxels are modeled as independent geometric extensions of the nodes to which they are attached.

The desired node displacements of the feet presented in figure \ref{fig:fig5} describe a gait wherein the robot takes a forward step with the right front foot and then the left front foot. At alternating points throughout the actuation sequence, the front feet take turns lifting while the back feet shift between the heel and toe of the foot and displace in the forward direction. Backward steps can be achieved by running the foot positions in reverse. \cite{spaeth_neuromorphic_2020}

A highlight of the POM approach is that in a single pass it can both compute the inverse kinematics to designate the actuation sequence as well as a sort of ``lateral kinematics'', equivalent to a composition of inverse and forward kinematics. This process predicts the motion of nodes which can be experimentally validated.

\begin{figure*}
    \centering
    \includegraphics[scale=0.4]{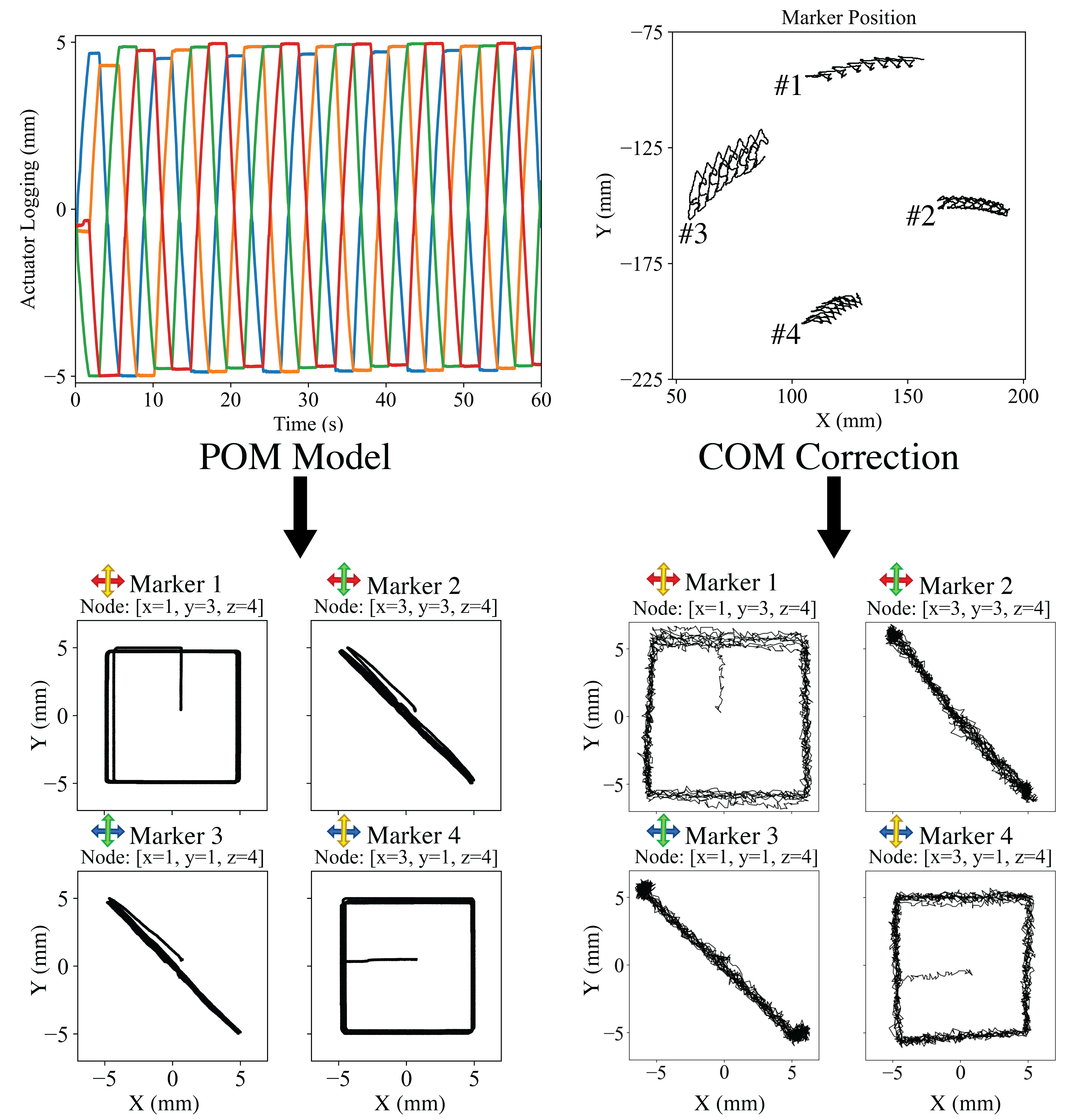}
    \caption{The actuator logging data is run through the POM model to estimate the displacement of the markers. The marker positions from the motion capture data are corrected for dynamic effects to estimate the displacements of the markers.}
    \label{fig:fig7}
\end{figure*}

\subsection{Robotic Experiment}

A physical realization of the designed voxel robot with 3D printed back feet and ``sticky'' front feet performs the specified gait on a smooth surface as shown in figure \ref{fig:fig5}.
Actuation is performed by four linear actuators which vary the total distance between attached nodes by $\pm10$~mm, displacing each by $\pm5$~mm. Details about the design and fabrication of the feet, voxel material properties, and linear actuators are provided in the supplementary information (SI).  

To track the position of the robot through time as it carries out its gait sequence, a 2D motion capture (mocap) system was set up. Four purple markers were placed on the nodes specified as end effectors to track. In addition to mocap data, actuator positions were also recorded via on-robot datalogging for comparison with model results.

To compare the observed mocap data to the predictions of the POM model, a preprocessing algorithm is employed which transforms the recorded coordinates $\vec m^L_i(t)$ of the mocap markers in the stationary lab frame to the corresponding coordinates $\vec m^R_i(t)$ in a robot-centered frame consistent with the model output. This transformation is necessary because the POM model does not address the contact dynamics that would be required to simulate the gait directly in the global frame.
First, the mean position of the four markers $\bar{\vec m}^L(t) = \frac14 \sum_{i=1}^4 \vec m^L_i(t)$ is subtracted, as this represents motion of the center of mass; then, the marker positions are transformed to polar coordinates $(r_i(t),\theta_i(t))$, and their common rotation $\bar\theta(t)$ is likewise subtracted to account for twisting and turning of the body. Then, the positions are converted back to Cartesian coordinates $\vec m^R_i(t)$, and the amplitude is rescaled according to the variance in marker position before and after the transformation. Each step of this correction is illustrated in the SI.
 
Figure \ref{fig:fig7} shows the data from the actuation logging and from the mocap system tracking the position of the markers in time. Here, the POM model is used for forward kinematics. The actuator data is plugged in as the actuation sequence to the POM model, along with the control and end effector node parameters, and produces the same node displacements as predicted in figure \ref{fig:fig6}, but with slight noise. The mocap data of the purple markers show that the robot turns while locomoting, which could be due to systematic asymmetries like voxel fatigue, actuator extension and contact dynamics. After transformation into the body frame, the markers follow the same shape as the predicted kinematic template.

\section{Conclusion} 
We have introduced a modeling method for flexible voxel robots which directly relates the positions of ``control'' (input) and ``end-effector'' (output) nodes, significantly reducing model space. The resulting kinematic templates can be used to design the robot and its movement. A locomotive voxel robot was designed using this framework and the approach was validated with physical experiments. End effector trajectories are predicted with surprising accuracy considering the simplicity of the model.
The validity of the POM model on voxel-based structures has exciting implications for modeling flexible robots. This POM model can be used to quickly design end effector trajectories for locomotive gaits over specific terrain or to achieve other robotic tasks, like grasping or object manipulation. To improve the robot's design and motion, an optimization algorithm could minimize the number of voxels and actuators in a structure, similar to work that has been done to computationally design mechanical characters whose end effectors follow user-defined trajectories \cite{coros_computational_2013}. To correct for systematic asymmetries, like turning or voxel fatigue, a calibration step could be included, where data collected from the movement of a voxel robot is fed into a trajectory optimization based on locomotive modeling, as proposed by Bittner et al. for planar serpent-like swimmers \cite{bittner_geometrically_2018}. 

%%%%%%%%%%%%%%%%%%%%%%%%%%%%%%%%%%%%%%%%%%%%%%%%%%%%%%%%%%%%%%%%%%%%%%%%%%

\section*{Acknowledgements} 

The authors would like to acknowledge the technical support of the
Braingeneers research group as well as a donation made possible by
Eric and Wendy Schmidt by recommendation of the Schmidt Futures
program.

\section*{Author Disclosure Statement}
The authors declare no conflicts of interest.

%%%%%%%%%%%%%%%%%%%%%%%%%%%%%%%%%%%%%%%%%%%%%%%%%%%%%%%%%%%%%%%%%%%%%%%%%%

\bibliographystyle{ieeetr}
\bibliography{TebyaniMainDocument.bib}
 
\end{document}